\documentclass[conference]{IEEEtran}
\IEEEoverridecommandlockouts
\pdfoutput=1

\usepackage{cite}
\usepackage{amsmath,amssymb,amsfonts}
\usepackage{algorithmic}
\usepackage{graphicx}
\usepackage[colorlinks]{hyperref}
\usepackage{newtxtext,newtxmath}
\usepackage{xcolor}
\usepackage{bm} 
\usepackage{upgreek}
\usepackage{adjustbox}

\def\BibTeX{{\rm B\kern-.05em{\sc i\kern-.025em b}\kern-.08em
    T\kern-.1667em\lower.7ex\hbox{E}\kern-.125emX}}
\begin{document}

\title{3D Scene Understanding at Urban Intersection using Stereo Vision and Digital Map\\
}

\author{\IEEEauthorblockN{Prarthana Bhattacharyya, Yanlei Gu, Jiali Bao, Xu Liu and Shunsuke Kamijo}
\IEEEauthorblockA{Graduate School of Information Science and Technology \\
The University of Tokyo\\
Tokyo, Japan \\
email: prarthana@kmj.iis.u-tokyo.ac.jp and kamijo@iis.u-tokyo.ac.jp}
}

\maketitle
\date{} 

\begin{abstract}
The driving behavior at urban intersections is very complex. It is thus crucial for autonomous vehicles to comprehensively understand challenging urban traffic scenes in order to navigate intersections and prevent accidents. In this paper, we introduce a stereo vision and 3D digital map based approach to spatially and temporally analyze the traffic situation at urban intersections. Stereo vision is used to detect, classify and track obstacles, while a 3D digital map is used to improve ego-localization and provide context in terms of road-layout information. A probabilistic approach that temporally integrates these geometric, semantic, dynamic and contextual cues is presented. We qualitatively and quantitatively evaluate our proposed technique on real traffic data collected at an urban canyon in Tokyo to demonstrate the efficacy of the system in providing comprehensive awareness of the traffic surroundings.
\end{abstract}


\section{Introduction}
Autonomous driving has gathered attention and tremendous interest in the past few decades, with their implementation in commercial cars seeming imminent. Generally, autonomous driving on highways has been demonstrably successful till now. But urban environments because of their complexity, still pose a challenging problem. Challenges include narrow lanes, sharp turns, congested intersections, obstacles, occlusions, blocked streets, parked vehicles, pedestrians, bicyclists and other moving vehicles. Traffic intersections are particularly crucial in this regard. Intersections are called ‘accident-hot-spots’ since misjudging the speed or intent of the surrounding vehicles can easily lead to disastrous collisions. Thus in order to ensure safe operation an autonomous vehicle should be able to continuously and reliably perceive its environment from its sensory inputs. For this purpose, it is evident that it is not just enough to detect the surrounding obstacles across each time step in isolation. The scene has to also be understood by the driving system in reference to the road-lane structure, ego-position, temporal context, as well as the driving task to be performed. 
\par \textbf{Contribution:} To achieve this aim, a vision and map-based approach is proposed in this paper to comprehensively understand the traffic situation at urban intersections. The surrounding traffic participants are detected, tracked, lane localized, while their spatial orientation and temporal behavior are integrated with the road-lane structure. 

\begin{figure}[t]
\centerline{\includegraphics[width=0.5\textwidth]{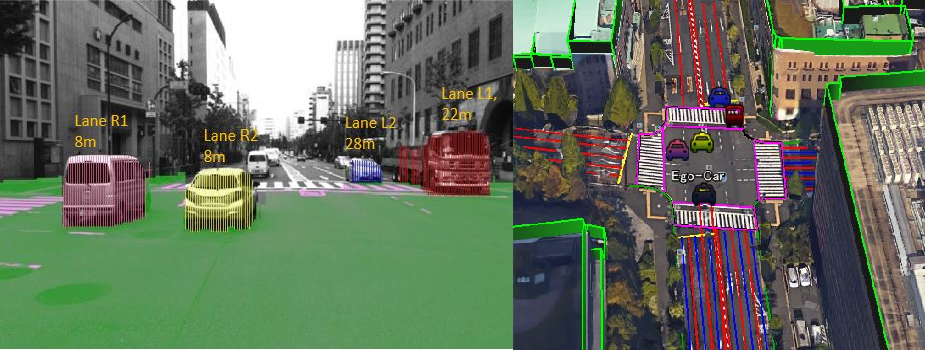}}
\caption{Understanding a typical urban traffic scene}
\label{intro1}
\vspace{-0.4cm}
\end{figure}

\par
\begin{figure}[t]
\centerline{\includegraphics[width=0.46\textwidth]{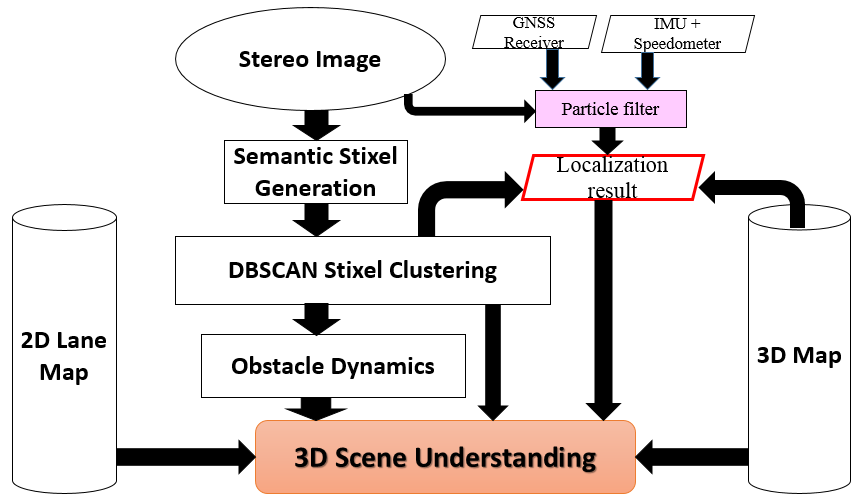}}
\caption{Flowchart of the Proposed Approach}
\label{flowchart}
\vspace{-0.2cm}
\end{figure}

 Fig. \ref{intro1} shows a typical example of traffic scene understanding at an urban intersection in Tokyo. The figure illustrates stereo-based free-space and vehicle detection and integration of their trajectory with the road structure information included in the 3D digital map of the surrounding. The corresponding flow diagram of the approach is shown in Fig. \ref{flowchart}.  The stereo camera input images are utilized to generate Semantic Stixels \cite{semanticstixels}, which is basically a way to compactly represent the obstacles present in a 3D scene with rectangles. An initial ego-position estimate is obtained using \cite{gu-san}, which is then refined by matching 3D building map data with the 3D stereo input evidence. Map matching also produces a heading direction estimate of the ego-vehicle, which is important to accurately localize surrounding vehicle trajectories. Once the obstacles have been identified and clustered, their dynamics are measured. Finally the Semantic Stixels clusters and their dynamic cues are probabilistically integrated with the map structure and ego-context to provide a 3D understanding of the traffic scene.

\section{Related Work}
Recently, many research works have utilized vision systems for comprehensive traffic situational awareness. A method proposed in \cite{guo-leader} detects and tracks surrounding vehicles while assigning them to their corresponding lanes, and also identifies a leader vehicle which is subsequently used for path planning. However this approach does not make special allowances for complex urban intersections as illustrated in Fig. \ref{intro1}. Another mid-level scene understanding platform is provided by \cite{stixelworld}, which uses stereo vision and models and tracks obstacles as rectangles with a fixed pixel width. Since they can be clustered together and tracked to impart the notion of objects, applications like \cite{roundabout} uses this approach to recognize dangerous situations at roundabouts.

\par However, it cannot reason about the scene comprehensively owing to a lack of context. An approach in \cite{geiger-sceneunderstanding} discusses the scene understanding problem from the point of using visual cues like semantic labels, scene flow and occupancy grids to infer the scene geometry and traffic activities, but do not accommodate prior map knowledge into their model. A generative model is proposed in \cite{modeltraffic} to reason about high-level scene semantics by stressing that there are limited, finite type of traffic patterns in the real-world situations and they can be learned. But the parameters learnt for a particular intersection may not be easily translatable to other intersections. Our approach builds on an extension of \cite{semanticstixels} and incorporates prior map structure to reason effectively about the 3D scene.

\section{Framework Of The Scene Understanding Approach}
We formulate an approach to probabilistically understand the 3D traffic scene at an intersection. The semantic and geometric cues are obtained from stereo disparity generation algorithms and deep learning based methods. The contextual cues comprise of ego-position and heading direction estimation which are facilitated by the particle filter based integration of various input sensor data. The dynamic cues are obtained from optical flow estimation techniques. These input evidences are discussed in Section (III-A). Section (III-B) provides the framework to fuse these measurements in order to gain a higher-level understanding of the scene in terms of spatial orientation and temporal behavior of the surrounding traffic participants.

\subsection{Input Evidence}
\subsubsection{Geometric and Semantic Cues of Obstacles}
\begin{figure}[b]
\vspace{-0.4cm}
\centerline{\includegraphics[width=0.46\textwidth]{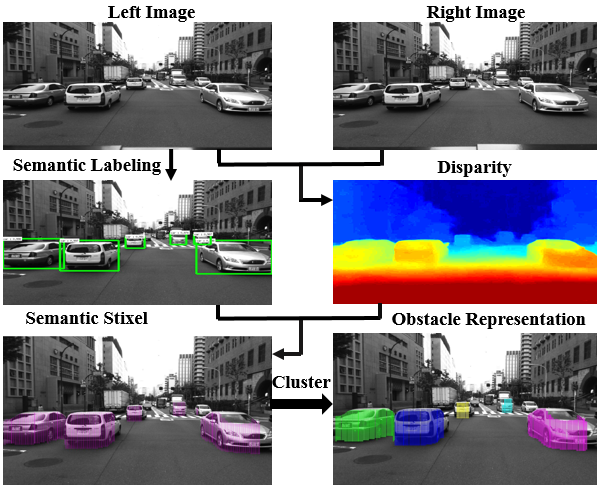}}
\caption{Integration of Semantic and Geometric Cues for Obstacle Detection}
\label{obstacle}
\end{figure}

Firstly the dense disparity images are estimated using DispNet \cite{dispnet}. In the next step, a state-of-the-art and publicly available region-based fully convolutional network (R-FCN) \cite{fasterrcnn} is used to generate pixel level probability scores for different semantic labels. To integrate the geometric disparity cues with semantic labels, a scene model presented in \cite{semanticstixels} is used to produce a set of Semantic Stixels $\textbf{S}_t$ at time-step $t$. A single Stixel $\textbf{s} \in \textbf{S}_t$ is defined by a five-dimensional vector $\textbf{s}=[u, v_b, v_t, d, l]$ . Here, $u$ is the image column and $v_b$ and $v_t$ mark the base and top point of the Stixel in image coordinates. The disparity value of the Stixel is $d$ and semantic class is $l$. Finally, the Semantic Stixels are grouped together in order to join every Stixel with a similar depth and semantic class into the same obstacle. Density-based spatial clustering of applications with noise (DBSCAN) \cite{directionaldbscan} algorithm is chosen by our approach for Stixel clustering since it does not require the predetermination of the number of clusters and can discover clusters with arbitrary shapes. This process assigns a cluster-id $k \in {1, ...,C}$ to each Stixel in the current scene, where $C$ is the number of obstacles at time $t$. An obstacle $\textbf{o} \in \textbf{O}_t$ contains a set of Stixels with the same cluster-id and is defined as $\textbf{o} = [\{s_i\}:$ cluster-id$(i)=k]$. The process of integrating semantic and geometric cues from stereo image input data by Semantic Stixels and clustering them to detect obstacles is illustrated in Fig. \ref{obstacle}.

\subsubsection{Vehicle Self-Localization and Context}
Accurate vehicle self-localization extremely important for scene understanding, and is the key to motion planning and vehicle cooperation. Positioning by Global Navigation Satellite Systems (GNSS) suffer from NLOS propagation and multi-path effects in the urban canyon, while inertial sensors increasingly drift with time. An integrated self-localization system, comprising of GNSS receivers, onboard-cameras and inertial sensors, is proposed in \cite{gu-san} for challenging urban city scenario. This method is modified in this work to include heading-direction correction of the ego-vehicle.
In this paper, there are four main sources of positioning, namely Global Navigation Satellite System (GNSS), Inertial Navigation Sensor (INS), stereo-vision and 3D building map. The 3D building map construction has been discussed in \cite{gu-san}. 

\begin{figure*}[t]
\centerline{\includegraphics[height=7cm, width=0.8\textwidth]{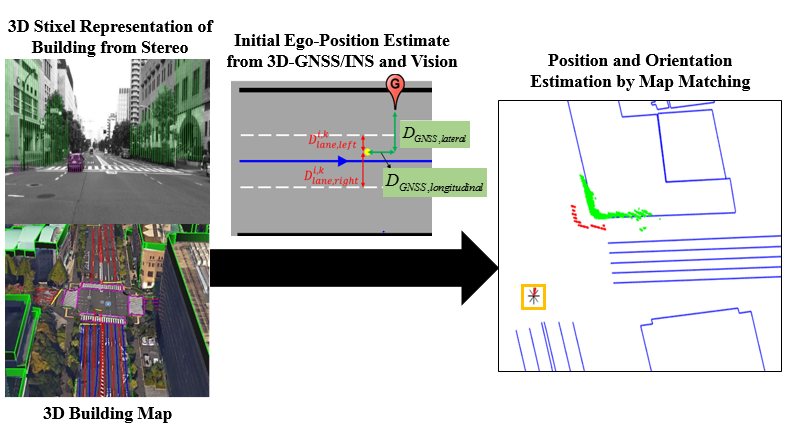}}
\caption{Estimation of ego-position and orientation. Initial ego-position estimate is found by particle filtering. 3D Building matching calculates orientation.}
\label{localization}
\vspace{-0.2cm}
\end{figure*}

The construction of the 3D map requires the 2-dimensional building footprint, which is provided by Japan Geospatial Information authority, and the Digital Surface Model (DSM) data, acquired from the Aero Asahi Corporation. The height information of the building is included in the DSM data. The 2D map on the other hand, is generated from high resolution aerial images provided by NTT-geospace. Particle filtering is used to integrate multiple sensor information from GNSS, INSS, vision and 3D map. 
\par INS describes motion of vehicle via the velocity and the heading direction. This information is used for particle propagation in the fusion algorithm. GNSS gives global localization measurement, which can estimate probability of particles. Vision based lane detection perceives the relative distance from the center of vehicle to left white line and right white line. This distance is used to refine the weight of the particles. 
\par In the particle filter, the system state is represented through a set of samples in the inertial frame $\{\textbf{x}_t=(x^t_{\text{north}}, x^t_{\text{east}})\}^{j=1,..n}$. Suppose that a set of $n$ random samples from the posterior probability distribution function $p(\textbf{G}_{t-1}, \textbf{V}_{t-1}|\textbf{x}_t^j)$ is available. GNSS positioning result is  $\textbf{G}_{t-1}$ and $\textbf{V}_{t-1}$ is the lane detection result at time $t-1$ respectively. Then the weighted average of all particles decides the localization result $\textbf{x}_{t-1}$ for the time $t-1$.
The particle weights are ascertained by equations \ref{eq:1}-\ref{eq:2} and shown in Fig. \ref{localization}, where $\mathcal{N}$ represents a normal distribution. $\sigma^2_{lane}$ and $\sigma^2_{GNSS}$ are empirically chosen. 
\begin{equation}
  p(\textbf{V}_{t}|\textbf{x}_t^j) = \mathcal{N}(D_{t, \text{left}}^j, \sigma^2_{\text{lane}}) \cdot \mathcal{N}(D_{t, \text{right}}^j, \sigma^2_{\text{lane}})  \label{eq:1} 
\end{equation}
\begin{equation}
    \begin{array}{c}
    p(\textbf{G}_{\text{lateral}}|\textbf{x}_t^j) = \mathcal{N}(D_{\text{GNSS}, \text{lateral}}^t, \sigma^2_{\text{GNSS}}) \\  
     p(\textbf{G}_{\text{longitudinal}}|\textbf{x}_t^j) = \mathcal{N}(D_{\text{GNSS}, \text{longitudinal}}^t, \sigma^2_{\text{GNSS}})
  \end{array} \label{eq:2}
\end{equation}

The joint posterior probability from which samples are drawn is represented as equation \ref{eq:3} where, is the credibility for GNSS measurement along the lateral direction. With the particle states and positioning result at $t-1$, the particle filter will exclude low-weighted particles, and recursively estimate the localization result $\textbf{x}_{t}$ for the ego-vehicle at time $t$.
\begin{equation}
    \begin{array}{c}
    p(\textbf{G}_{t-1}, \textbf{V}_{t-1}|\textbf{x}_t^j) =  
    \{\gamma \cdot  p(\textbf{G}_{t, \text{lateral}}|\textbf{x}_t^j) +
    \\
    (1-\gamma) \cdot  p(\textbf{V}_{t}|\textbf{x}_t^j)\} \cdot p(\textbf{G}_{t, \text{longitudinal}}|\textbf{x}_t^j) 
  \end{array} \label{eq:3}
\end{equation}

For estimating the heading direction $\theta$ of the ego-vehicle, Normal Distributions Transform (NDT) \cite{ndt} based map matching is used. Equation \ref{eq:4} represents the spatial matching of the building points from the digital map  (shown in green color in Fig. \ref{localization}) and 3D Stixel representations of buildings (shown in red color in Fig. \ref{localization}). The idea is to probabilistically align these two spatial vectors in order to recover the parameter $\theta$.
\begin{equation}
    \begin{pmatrix}
m_{\text{north}}\\
m_{\text{east}} 
\end{pmatrix} = \begin{pmatrix}
cos\theta & -sin\theta\\
sin\theta & cos\theta 
\end{pmatrix} \begin{pmatrix}
b_{\text{north}}\\
b_{\text{east}} 
\end{pmatrix} 
\label{eq:4}
\end{equation}
The mechanism of position and orientation estimation of the ego-vehicle with respect to the 3D digital map by using GNSS, INS, and stereo-vision information is illustrated in Fig. \ref{localization}. The black star-shaped symbol denotes the calculated 2D ego-position on the map and a red-dotted line emanating from it represents its heading angle. This provides contextual information for localizing other traffic participants on the map.

\subsubsection{Dynamic Cues of Traffic Participants} Optical flow provides strong cues for temporal scene understanding. In this paper, the popular Lucas-Kanade method \cite{opticalflow} is used to calculate the optical flow vectors $\textbf{U}_{t-1, t}$ for obstacles $\textbf{o} \in \textbf{O}_t$, which was obtained in Section (III-A-1).

\subsection{Integration}
This section presents a probabilistic fusion technique to temporally combine the semantic, geometric, contextual and dynamic cues and output the digital map position and tracked object-id for all surrounding traffic participants across different time-steps to produce a holistic understanding of the traffic scene. 
\\ The  position  state $\textbf{\'X}_t$ of obstacle $\textbf{o}$ with respect to the camera frame of reference is defined as $(\textbf{\'X}_{\text{north}}$, $\textbf{\'X}_{\text{east}})^T$. In order to obtain an analytic solution to the state, the state transition model is assumed to be linear-Gaussian and is given by equations \ref{eq:5}-\ref{eq:6}. The initial velocity of all objects is set to around 6 $m/s$ empirically. For the consecutive frames, $\textbf{B}_t = \textbf{B}_{t-1}$. The measurement model is non-linear and is given by \ref{eq:7}. The process and measurement noise vectors $\bm{\upomega}$ and $\bm{\upnu}$ are assumed to be Gaussian white noise.
\begin{equation}
    \textbf{\'X}_t = \textbf{A}_t\textbf{\'X}_{t-1} + \textbf{B}_t + \bm{\upomega}_t
\label{eq:5}
\end{equation}
\begin{equation}
    \textbf{A}_t = \begin{bmatrix}
1 & 0 \\
0 & 1 
\end{bmatrix} , \textbf{B}_t = \begin{bmatrix}
\Delta t \cdot v_{\text{north}, t}  \\
\Delta t \cdot v_{\text{east}, t}
\end{bmatrix}
\label{eq:6}
\end{equation}
\begin{equation}
    p(\textbf{Z}_t|\textbf{X}_t) = p(\textbf{O}_t,  \textbf{U}_{t-1, t},  \textbf{x}_t|\textbf{X}_t)
\label{eq:7}
\end{equation}

An obstacle $\textbf{o} \in \textbf{O}_t$ at time-step $t$, is assigned the measured image coordinates $(u_{\text{center}}, v_{T, \text{center}})_t$ and disparity value $d_{\text{center}}$ from its Stixel cluster. Equation \ref{eq:8} relates the measured obstacle positions at time step $t$ and $t-1$ with the optical flow.
\begin{equation}
\label{eq:8}
(u_{\text{center}}, v_{T, \text{center}})_t = (u_{\text{center}}, v_{T, \text{center}})_{t-1} + \textbf{U}_{t-1, t} 
\end{equation}
Assuming a pin-hole camera model with baseline $b’$ and focal length $f_u$ \cite{uvdisparity}, the equation for the measurement update at time $t$ is:
\begin{equation}
\begin{bmatrix}
u_{\text{center}}  \\
d_{\text{center}}
\end{bmatrix}_t = 
\begin{pmatrix}
{\frac{1}{X'_\text{north}} \begin{bmatrix}
X'_{\text{east}} \cdot f_u  \\
b' \cdot f_u
\end{bmatrix}} 
\end{pmatrix}_t  + \bm{\upnu}
\end{equation}

An Extended Kalman Filter (EKF) \cite{6dvision} is used to obtain $\textbf{\'X}_t$, the filtered position of $\textbf{o} \in \textbf{O}_t$. The final position of the surrounding traffic participants on the digital map is obtained according to the following equation:
\begin{equation}
    \textbf{\'X}_t = \textbf{R}_y(\theta)\textbf{\'X}_t + \textbf{x}_t
\end{equation}
Here $\textbf{R}_y$ represents the rotation matrix around $y$-axis. The temporal association of observation  $\textbf{Z}_t$ to an object $\textbf{o}$ at time $t-1$ is based on scoring the Euclidean distance to the prediction.

\section{Experimental Evaluation }
\begin{figure}[b]
\centerline{\includegraphics[width=0.46\textwidth]{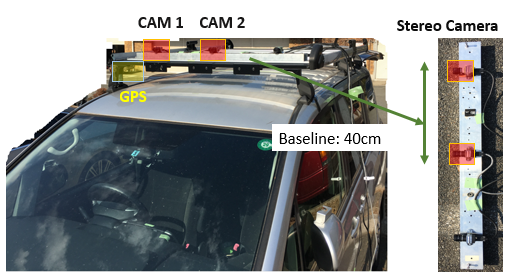}}
\caption{Experimental setup to collect real traffic data}
\label{setup}
\end{figure}
The purpose of the experiment is to evaluate our presented approach on real traffic data collected at an urban intersection in Tokyo. In order to evaluate the 3D scene understanding achieved by our proposed technique, we select single-object based tasks as well as a final total scene understanding task that integrates everything. Average accuracy of object detection, semantic labeling, and self-localization are some popular tasks considered for evaluation. We also measure motion states, trajectory, lane information and positioning information of 

\begin{table}[t]
 \caption{QUANTITATIVE EVALUATION}
  \label{table1}
  \centering
   \begin{adjustbox}{max width=0.48\textwidth}
   
   \begin{tabular}{c|c|c|c}
      \hline 
      \multicolumn{1}{c|}{\textbf{Object Detection}} & \multicolumn{1}{c|}{\textbf{Detection Rate}} & \multicolumn{1}{c|}{\textbf{False Positive}} & \multicolumn{1}{c}{{\textbf{\shortstack{Frames with\\ False Positive}}}} \\ 
      \hline \hline
      \shortstack{Stixel World \cite{stixelworld} + \\ DBSCAN Clustering} & 87.2\%  &  21.6\%  & 114 \\ 
      \hline
      Our Method   &  \textbf{92.4\%} & \textbf{0.05\%}  & \textbf{27} \\ 
      \hline
    \end{tabular}
  \end{adjustbox}
\end{table}

\begin{table}[h]
  \centering
   \begin{adjustbox}{max width=\textwidth}
   
   \begin{tabular}{c|c|c}
      \hline 
      \multicolumn{1}{c|}{\textbf{Object Tracking}} & \multicolumn{1}{c|}{\textbf{MT}} & \multicolumn{1}{c}{\textbf{ML}} \\ 
      \hline \hline
      Our Method   &  42.1\% & 10.3\%  \\ 
      \hline
    \end{tabular}
  \end{adjustbox}
\end{table}

\begin{table}[h!]
  \centering
   \begin{adjustbox}{max width=\textwidth}
   
   \begin{tabular}{c|c}
      \hline 
      \multicolumn{1}{c|}{\textbf{Self-Localization}} & \multicolumn{1}{c}{{\textbf{\shortstack{Lane-Localization Rate of \\ Surrounding Traffic}}}}  \\ 
      \hline \hline
      Without building matching \cite{gu-san} & 77.3\% \\ 
      \hline
      Our Method & \textbf{94\%} \\
      \hline
    \end{tabular}
  \end{adjustbox}
  \vspace{-0.4cm}
\end{table}

surrounding traffic participants from the ego-vehicle. The final task is to integrate both object detection and road map layout to recognize and localize all objects in the road structure across different time frames.

\subsection{Experimental Setup}
The experimental setup is shown in Fig. \ref{setup}. U-blox EVK-M8 GNSS model was used to receive the GPS signals and was mounted on top of the vehicle. Stereo camera is made up of two Point Grey monocular cameras with the baseline of 400mm and set on top of the vehicle facing towards the front. The baseline of the camera is flexible from 300mm to 900mm. For urban city scene, the baseline is set to be 400mm in order to detect both near and far objects. One point grey monocular camera is set inside the vehicle as front view camera to record the ground truth trajectory. CAN data and MEMS-gyroscope data are taken from inertial sensors installed in the vehicle. The stereo data obtained has resolution 15fps, 1024 $\times$ 768 pixel.

\begin{figure*}[t]
\centerline{\includegraphics[width=\textwidth]{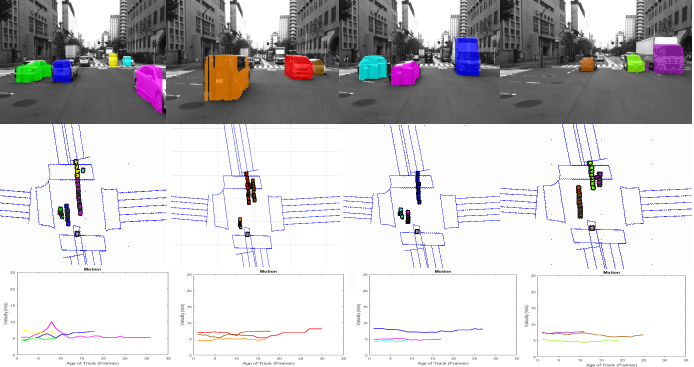}}
\caption{Qualitative results of the evaluated method. Integration of object segmentation and road map layout for spatial and temporal 3D scene understanding from ego-vehicle platform is demonstrated. The top row shows the obstacle semantic segmentation. The second row localizes the trajectory of the detected obstacles over tine on a map. The third row plots their velocities with respect to the age of their corresponding trajectories.}
\label{qual}
\vspace{0.46cm}
\end{figure*}

\subsection{Experimental Location}
The experiment is performed in Hitotsubashi area of Tokyo, where the density and height of buildings is typical for an urban canyon.

\subsection{Quantitative Evaluation}
In order to quantitatively evaluate our approach, an 800 frame video at the intersection is chosen, comprising of 80 vehicle sequences. Obstacle detection rate, surrounding vehicle localization rate and tracking metrics Mostly Tracked (MT) and Mostly Lost (ML) \cite{learningtotrack} are computed as shown in Table \ref{table1}. In the absence of exact ground truth trajectory, we calculate the lane-localization rate of the surrounding obstacles detected in order to estimate the ego-positioning accuracy.
\par We compare our results to two techniques: results of the DBSCAN clustered Stixel World approach described in \cite{stixelworld} and the particle filter based positioning method described in \cite{gu-san} without 3D map matching.
\par Our results show that object detection based on clustering semantic stixels together with digital map reduces the frames with false positives (FP) drastically. Our comparison also shows that the use of digital map for heading direction estimation helps to reduce the lane-localization error of surrounding vehicles, making the understanding of the system more robust. Tracking results are also presented to describe the temporal correspondence accuracy. We conclude that 3D digital maps help to considerably enhance the scene understanding capabilities of a purely vision based system.

\subsection{Qualitative Evaluation}
Fig. \ref{qual} illustrates the qualitative results of the proposed method on four example sequences. The top row shows the Semantic Stixel segmentation results of the input images. Accurate and compact representation of the surrounding traffic participants is obtained. The second row shows the integrated spatial and temporal states of the detected obstacles, while their corresponding trajectories are localized on the digital map. The self-localized ego-vehicle is depicted in black. For most sequences the vehicles are assigned to the correct lanes. The vehicle-to-trajectory correspondences are correctly maintained over time. The measured motion state of the surrounding traffic with respect to the age of the trajectory is shown in the third column. This velocity profile is important for the ego-vehicle in order to distinguish between passing and turning cars at the intersection. Overall, an effective perception of the surrounding traffic environment is achieved by this approach.

\section{Conclusion}
In this work, we presented a stereo vision and digital map based framework for robust self-localization and accurate 3D urban traffic scene perception. Additionally a probabilistic fusion of geometric, semantic, dynamic and contextual cues is presented to reason about the scene at a higher level and account for uncertainty in sensor information. Furthermore, precise heading direction estimation of ego- vehicle at turning- intersections and their influence on surrounding traffic localization is addressed. The proposed approach can be used at an urban intersection to answer questions such as: where the ego-vehicle is located on a given digital map; where the surrounding vehicles are located; which car is driving on which street; what their trajectory history is and what the current traffic states are.
\par Low-cost, close-to-production sensors are leveraged to tackle the challenging accident-prone urban intersection scenario. We quantitatively evaluate surrounding obstacle detection, positioning and temporal correspondence on real urban traffic data and achieve high performance. Qualitative evaluation depicts the integration of the detected obstacles with the road map layout and is shown to provide comprehensive situational awareness. In the future, we plan to use the measured velocity information and lane context of the detected traffic participants to probabilistically infer their intent and predict their maneuver.

\bibliographystyle{IEEEbib}
{\footnotesize
\bibliography{refs.bib}}
\end{document}